\documentclass{article}

    \usepackage[nonatbib, final]{neurips_2021}
\usepackage[utf8]{inputenc} 
\usepackage[T1]{fontenc}    
\usepackage{hyperref}       
\usepackage{url}            
\usepackage{booktabs}       
\usepackage{amsfonts}       
\usepackage{nicefrac}       
\usepackage{microtype}      
\usepackage{xcolor}         

\usepackage{amsmath}
\usepackage{graphicx}
\usepackage{algorithm}
\usepackage{algpseudocode}
\usepackage{float}
\usepackage{subfig}
\graphicspath{{}{figs/}}

\title{Achieving Low Complexity Neural Decoders via Iterative Pruning}

\author{%
  Vikrant Malik \\
  Indian Institute of Technology Kanpur, India \\
  \texttt{vikrant@iitk.ac.in}
  \And
  Rohan Ghosh and Mehul Motani \\
  National University of Singapore \\
  \texttt{rghosh92@gmail.com, motani@nus.edu.sg}
}

\begin{document}

\maketitle

\begin{abstract}
    The advancement of deep learning has led to the development of neural decoders for low latency communications. However, neural decoders can be very complex which can lead to increased computation and latency. We consider iterative pruning approaches (such as the lottery ticket hypothesis algorithm) to prune weights in neural decoders. Decoders with fewer number of weights can have lower latency and lower complexity while retaining the accuracy of the original model. This will make neural decoders more suitable for mobile and other edge devices with limited computational power.
    \textcolor{black}{We also propose \textit{semi-soft decision decoding} for neural decoders which can be used to improve the bit error rate performance of the pruned network.}
\end{abstract}

\section{Introduction}

Modern fifth generation (5G) communication systems require ultra-reliable low-latency communications (URLLC) \cite{sutton2019enabling}. This necessitates the use of low latency decoders which can infer the sent data from the received data quickly. \textcolor{black}{Deep learning based function approximation approaches provide a natural way to side-step many of the computational complexities involved in various sub-parts of the communication pipeline, enabling low-latency communications \cite{llnd, multiconcat, multilabel, o2017, zheng2020deepreceiver, nachmani2018deep, kim2018communication}. Such approaches could thus potentially find use in mobile and other low-end devices which have limited computational power.} 


\textcolor{black}{One such example of a deep learning based approach that has been explored is the use of neural networks for low-latency decoding \cite{llnd, multiconcat}.  Typically, neural decoders for $(n,k)$ codes can either be single-label ($2^k$ output nodes) or multi-label ($k$ or $n$ output nodes). Here, $k$ represents the number of message bits, and $n$ represents the number of transmitted codeword bits. Thus single-label decoders become computationally unfeasible with larger $k$, as opposed to multi-label decoders. However, as observed in \cite{doan2018neural}, even for multi-label decoders, the desired network complexity increases substantially with larger $k$, due to the larger dimensionality of the input space. As the primary goal of using these networks is to reduce computational complexity, there remains much scope for exploring ways to reduce complexity of these architectures while maintaining their decoding performance. }

\textcolor{black}{To address the issue of complexity, we explore a state-of-the-art approach for reducing neural network complexity while maintaining performance, namely the iterative network pruning approach proposed in \cite{LTH}, which uses the lottery ticket hypothesis (LTH). The lottery ticket hypothesis states that smaller sub-networks exist within the original network which can reach test accuracy comparable to the original model (or greater), when trained in isolation (in a similar number of iterations). In this paper, we apply the iterative pruning method in \cite{LTH} to neural decoders for Hamming and polar codes, and find that we can substantially reduce the complexity of neural decoders (up to 97\% pruning rates) while retaining their performance (test accuracy and Bit-Error Rate). Since the main purpose of the neural decoders is to correct errors introduced by the channel, we only consider neural decoders which predict the denoised codeword itself, i.e., using $n$ input and $n$ output nodes.\footnote{Note that for systematic codes, this is equivalent to a neural decoder that directly infers the message bits because the message bits can be directly inferred from the codeword bits.}  Additionally, we also propose a \textit{semi-soft decision decoding} approach for neural decoders, which enhances the performance of a neural decoder while minimally increasing computational complexity.}

\textcolor{black}{Our work in this paper provides a platform for (i) further exploration of complexity reduction approaches such as the iterative-pruning approach in \cite{LTH} for neural decoding, and (ii) also the use of computationally feasible extensions for neural decoders such as semi-soft decoding for improving decoding performance.}


\section{Background}


In a communication system the transmitter encodes a message of $k$ bits into a code word of length n. This codeword is then sent over an additive white Gaussian noise (AWGN) channel to the receiver where it decodes the received vector of length $n$ to find out the original codeword (or message). In the optimal decoding strategy, also called maximum likelihood (ML) decoding, the receiver chooses the codeword which is closest (in terms of Euclidean distance) to the noisy received vector. Since the total number of codewords is $2^k$, ML decoding needs to compare each received block to $2^k$ codewords, which grows exponentially with $k$. To overcome the exponential decoding complexity of soft decision decoding, neural decoders can be used (\cite{multilabel}) to quickly produce an estimate of the sent codeword, eliminating the need to compare the received vector with all the $2^k$ valid codewords.

\textcolor{black}{In this work, we also study one-shot pruning and the iterative pruning approach shown in \cite{LTH} to reduce the complexity of our neural decoders. One-shot pruning removes a fraction of the weights after one round of training. In contrast, iterative pruning involves recursively pruning a small proportion of the network weights at the end of each iteration of training, which is invoked via the early stopping criterion mentioned in \cite{LTH}. After the pruning step at the end of each training iteration, the rest of the un-pruned network weights are reset back to their initial values, after which the entire training and pruning process is repeated. It was shown in \cite{LTH} that such an iterative pruning method can yield heavily pruned network configurations which can perform similar or better than the original network.}

\section{System Model}

\subsection{Communication Model}

In the communication model, the transmitter encodes the message $m$ consisting of $k$ information bits, i.e., $m \in \{0, 1\}^k$, to a codeword consisting of $n$ coded bits, i.e., $c \in \{0, 1\}^n$. Each of the message bits can be either 0 or 1 with equal probability. Then, the transmitter modulates the codeword to get the modulated signal $s \in \{-1, +1\}^n$. These modulated bits are then sent over an AWGN channel to the receiver. \textcolor{black}{At the receiver, a noisy corrupted form of the codeword is received, i.e., $r=s+\epsilon$, where $\epsilon\in \mathbb{R}^n$ is the noise vector sampled from the multivariate Gaussian distribution $\mathcal{N}(0,\Sigma)$. For an AWGN channel, $\Sigma=\frac{N_0}{2}I_n$, where $I_n$ denotes the identity matrix of size $n\times n$ and $N_0$ is the power spectral density of the noise.}
The received vector $r$ is then fed into a neural network which outputs a prediction vector $p \in [0, 1]^n$. We denote the rate of this code by $R = k/n$. Moreover, the energy per message bit is given by $E_b$.

\subsection{Neural Decoder Architecture}
The neural decoder consists of $H$ hidden layers with $N_i$ neurons each for $i \in \{1, 2, \dots H\}$. The input and output layers consists of $n$ neurons each, where $n$ is the codeword length. We use the ReLu function $\sigma_{i} (x) = \max{(0, x)}$ as the activation function for our hidden layers. Moreover, the sigmoid function $\sigma_o (x) = \frac{1}{1 + \exp(-x)}$ is used as the activation function for the output layer. 
Here, the sigmoid functions restricts the output of the model to be between 0 and 1. which can be seen as the probability of the bit being 0 or 1 at one particular position. The model is trained with the loss function being the binary cross entropy function.
Table \ref{tab1} summarizes the parameters of the neural decoder.

\begin{table}[!t]
\caption{Parameters of the neural decoder model}
\label{tab1}
\begin{center}
\begin{tabular}{ |c|c|c| } 
 \hline
 \textbf{Layer} & \textbf{Activation Function} \\ 
 \hline
 Input Layer ($L_{input}$) & None \\ 
 \hline
 Hidden Layer ($L_i$ for $i \in {1, 2, \dots H}$) & $\sigma_i(x)$ \\ 
 \hline
 Output Layer ($L_{output}$)  & $\sigma_o(x)$ \\ 
 \hline
\end{tabular}
\end{center}
\end{table}

\section{Decoding Techniques}

\subsection{Hard Decision Decoding}
To output a binary codeword, we use threshold based decoding, with threshold $\tau = 0.5$, on the neural decoder output of length $n$. That is, for each bit position $i \in \{1, 2, \dots n\}$, we decide that the codeword bit at position $i$ is 0 if position $i$ of the output of the neural network is less than $\tau$, else we decide that the bit at position $i$ is 1. 

\subsection{Semi Soft Decision Decoding}
To improve the performance of the pruned network, we use soft decision decoding on a fixed number of the predicted codeword bits and do a hard decision (threshold) decoding on the rest. This approach, termed semi-soft decision decoding, is inspired by Chase decoding \cite{chase1972class} and its variants \cite{chase}. 
First, for a received vector of length $n$ and for a fixed number $b \in {0, 1, \dots n}$, we find the $b$ "least confident bit positions" (which are closest to 0.5) of the output of the neural network. \textcolor{black}{Then, we do threshold decoding on the rest of the bits and keep these $b$ bit positions the same. Let's denote the prediction vector at this stage by $c_{ref}$. We now fill the remaining $b$ bits with all possible combinations of 0s and 1s and obtain a list of $2^b$ vectors. Now, using a hash table which consists of all the $2^k$ valid codewords, we can check for each of the $2^b$ possibilities, in $O(1)$ time, whether they are present in the hash table. Having obtained all the valid codewords that match $c_{ref}$ in the remaining $n - b$ positions, we can compare the euclidean distance of $c_{ref}$ with each of the valid codewords that we just obtained and chose the codeword that is closest to $c_{ref}$ to get the final estimated codeword as $\hat{c}$. Note that the maximum number of valid codewords that we can obtain using the hash table is $2^b$. Hence, we can reduce the number of codewords ($\leq 2^b$ instead of $2^k$) with which we have to compare the distance of the output of the neural network. In the case when there are no valid codewords for a given $p_{n-b}$, we simply do a hard decision decoding the $b$ bits as discussed in Section 4.1.} The complexity of this algorithm $O(2^b)$ can be controlled by the receiver by specifying the $b$.



\section{Experiments}
\textcolor{black}{In this section, we conduct a series of experiments with neural decoders for Hamming (7,4) \cite{lin2001error} and Polar (16,8) codes \cite{polar}.} \textcolor{black}{In the case of Hamming Codes, the BER is the fraction of message bits that are in error. For Polar Codes, we calculate the BER by finding the fraction of bits that are in error in the received codeword. The actual BER, which is the fraction of bits that are in error in the received message, may be slightly better owing to the fact that it is possible to have 0 errors in the estimated message even if there are some errors in the estimated codeword.} \textcolor{black}{In both cases, we first test the iterative pruning approach based on the LTH, and then test our proposed semi-soft decoding approach on the neural decoders. Unless mentioned otherwise, the training and testing SNR is 0 dB for Hamming Codes and 2 dB for Polar Codes. For what follows, we denote the optimal max-likelihood soft decision decoding approach as ML decoding, and the iterative pruning method is simply denoted as the LTH approach. }


\subsection{Hamming (7,4) Codes}

\begin{figure}[!t]
    \centering
    \includegraphics[width = 0.92\linewidth]{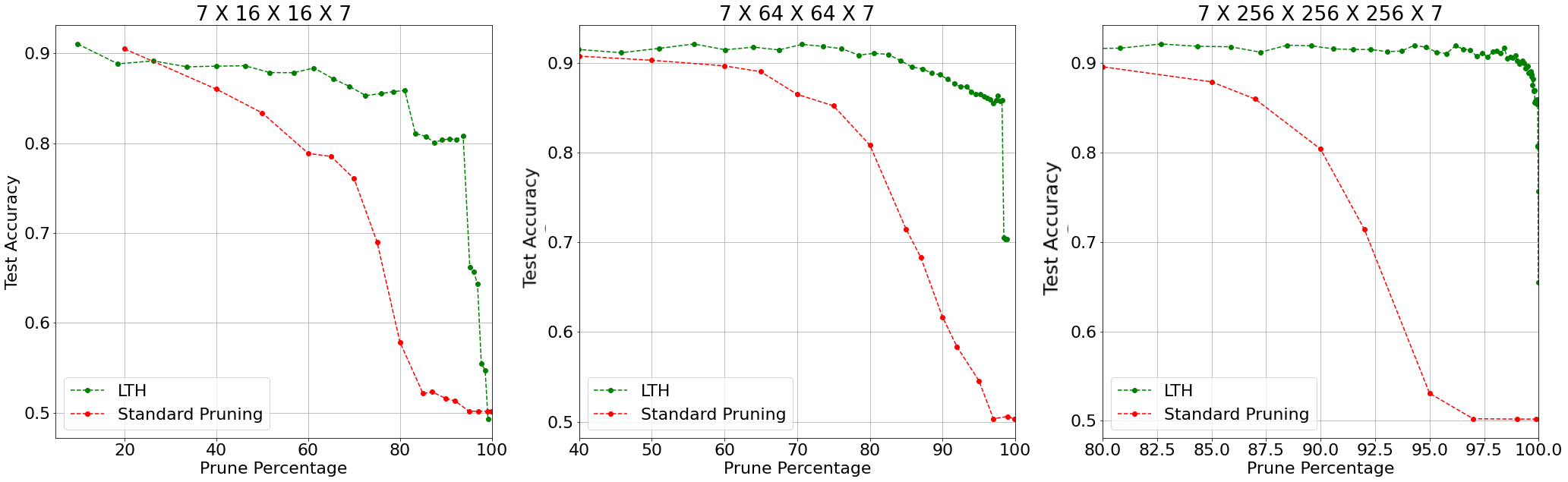}
    \caption{Comparison of Test Accuracy at various stages of pruning for Hamming (7, 4) codes (trained at 0 dB SNR). Network configurations are shown at the top of the plot in the following format: $L_{input} \times L_i \times L_{output}$ for $i \in {1, 2, \dots H}$.}
    \label{fig:2}
\end{figure}

\begin{figure}[!t]
    \centering
    \includegraphics[width = 0.92\linewidth]{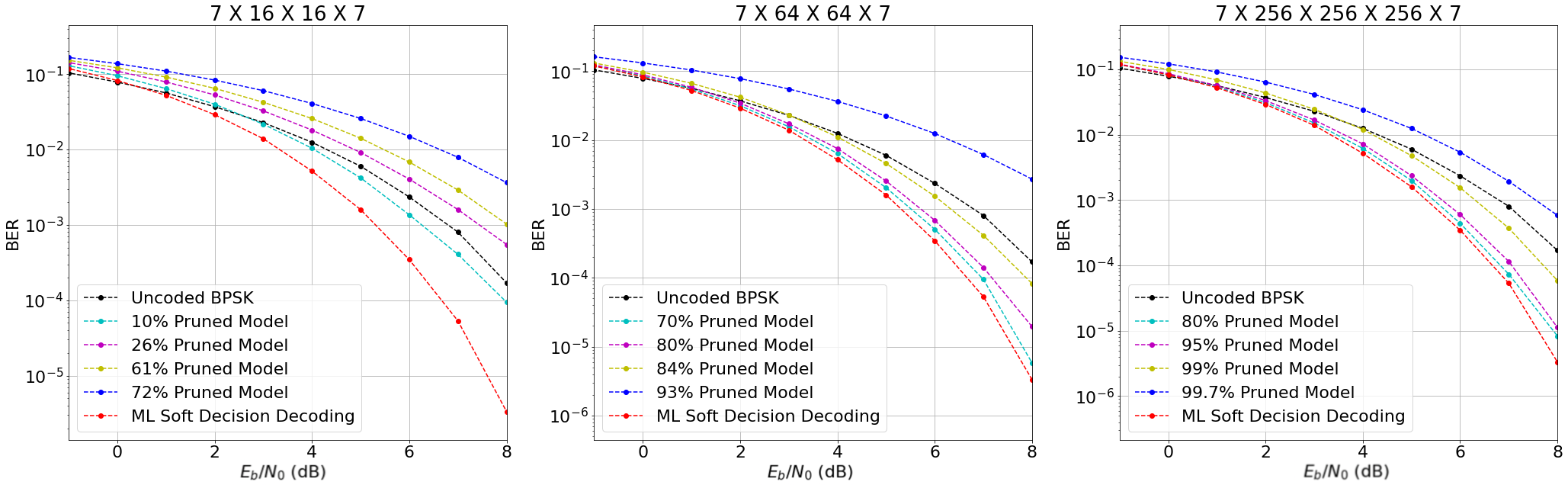}
    \caption{BER for neural decoders of various complexities for Hamming (7, 4) codes (trained at 0 dB SNR). Network configurations are shown at the top of the plot in the following format: $L_{input} \times L_i \times L_{output}$ for $i \in {1, 2, \dots H}$}
    \label{fig:3}
\end{figure}

\textcolor{black}{First, we test the LTH and the standard one-shot pruning approaches with neural decoders for Hamming (7,4) codes. Figure~\ref{fig:2} shows how the test accuracy changes in response to increasing levels of pruning, for three network configurations having varying levels of complexity. The network configurations are written at the top of the plots in the following format: $L_{input} \times L_i \times L_{output}$ for $i \in {1, 2, \dots H}$. We note that in all cases, the LTH outperforms standard pruning by a significant margin, while retaining the accuracy of the original model for a wide range of pruning. Figure~\ref{fig:3} shows the performance of the pruned network at various stages of pruning. By comparing Figure~\ref{fig:2} and Figure~\ref{fig:3}, we see that for the neural decoder to reach ML decoding level performance, the test accuracy in Figure \ref{fig:2} must be around 90\%. For the case of the decoder with the configuration of $L_i$ = 64 for $i \in {1, 2}$, we see that it can closely approximate the optimal ML Decoding when 80\% pruned. Interestingly, we consistently find that heavily pruned complex network configurations usually fare better than lightly pruned network configurations of less complexity.}

\begin{figure}[t!]
    \centering
    \includegraphics[width = 0.92\linewidth]{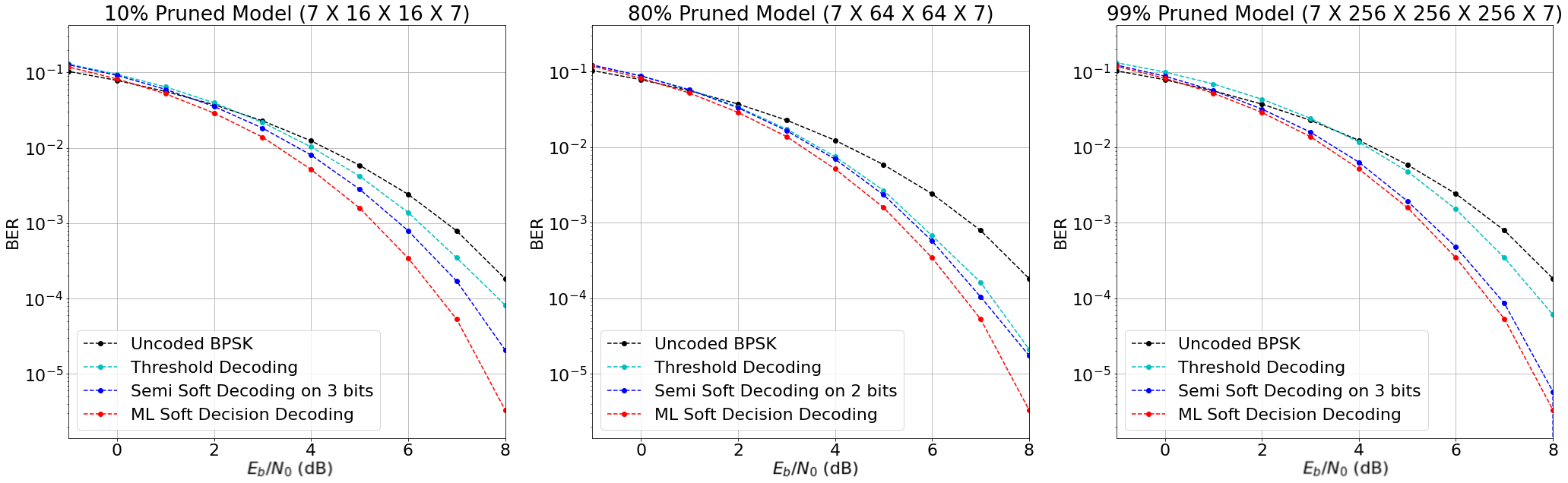}
    \caption{Effect of Semi-Soft Decision Decoding on the BER. Network configurations are shown at the top of the plot in the following format $L_{input} \times L_i \times L_{output}$ for $i \in {1, 2, \dots H}$}
    \label{fig:4}
\end{figure}

\textcolor{black}{
Next, in Figure ~\ref{fig:4}, we show the effect of the proposed semi-soft decision decoding method on the BER performances of pruned neural decoders. As can be seen from the figure, semi-soft decision decoding provides a noticeable improvement in the BER of the neural decoders even when $b$ is as small as 2 and 3. Moreover, we see a significant level of improvement for heavily pruned models, which originally have poor BER performance.  }


\vspace{1cm}

\newpage


\begin{figure}[t!]
\vspace{-4mm}
\centering
\quad  \subfloat[Test Accuracy (at 2 dB) vs Pruning Percentage.]{
    \includegraphics[width=0.4\textwidth]{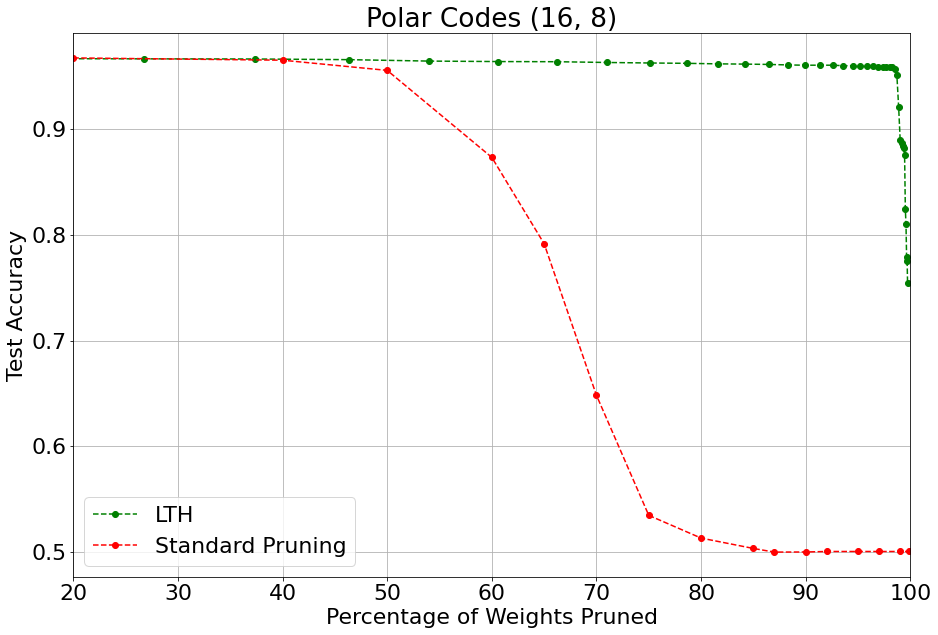}
    \label{fig:5a}
    }
    \quad
\subfloat[BER performance at various stages of pruning.]{
    \includegraphics[width=0.4\textwidth]{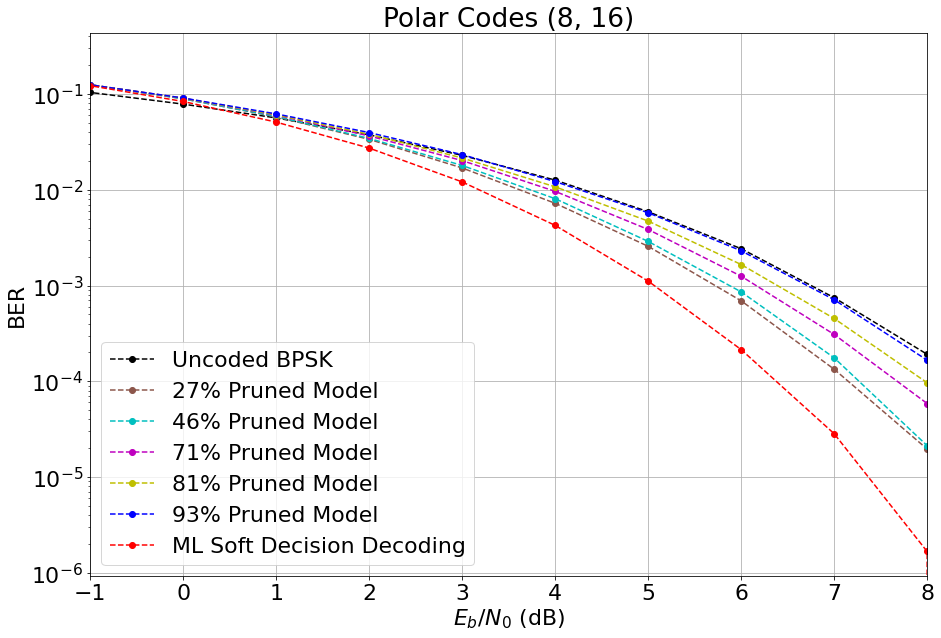}
    \label{fig:5b}
    }
    \newline
\subfloat[Effect of Semi-Soft Decision Decoding at 27\% Pruned Model]{
    \includegraphics[width=0.4\textwidth]{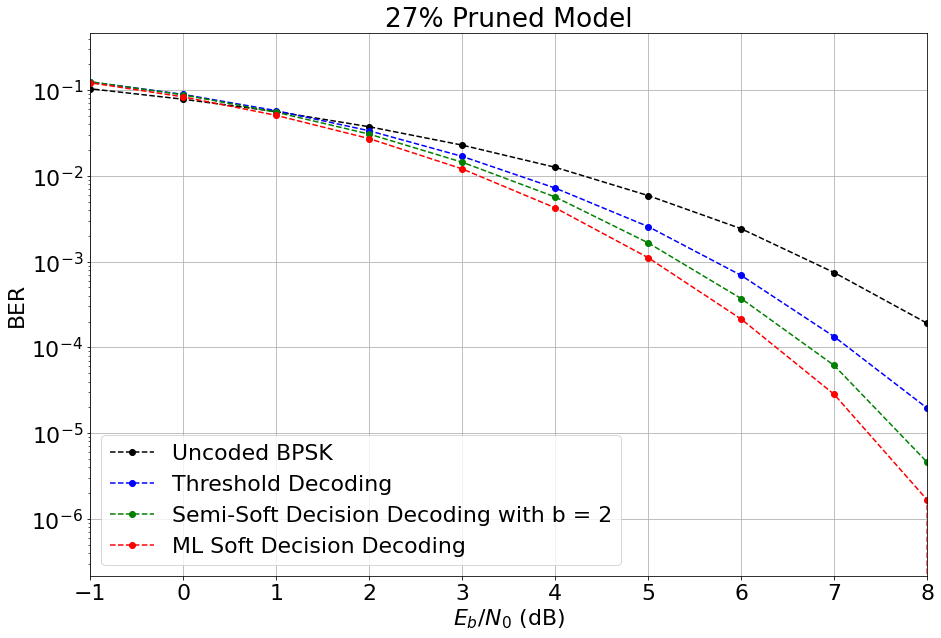}
    \label{fig:5c}
    }
    \quad
\subfloat[Effect of Semi-Soft Decision Decoding at 71\% Pruned Model]{
    \includegraphics[width=0.4\textwidth]{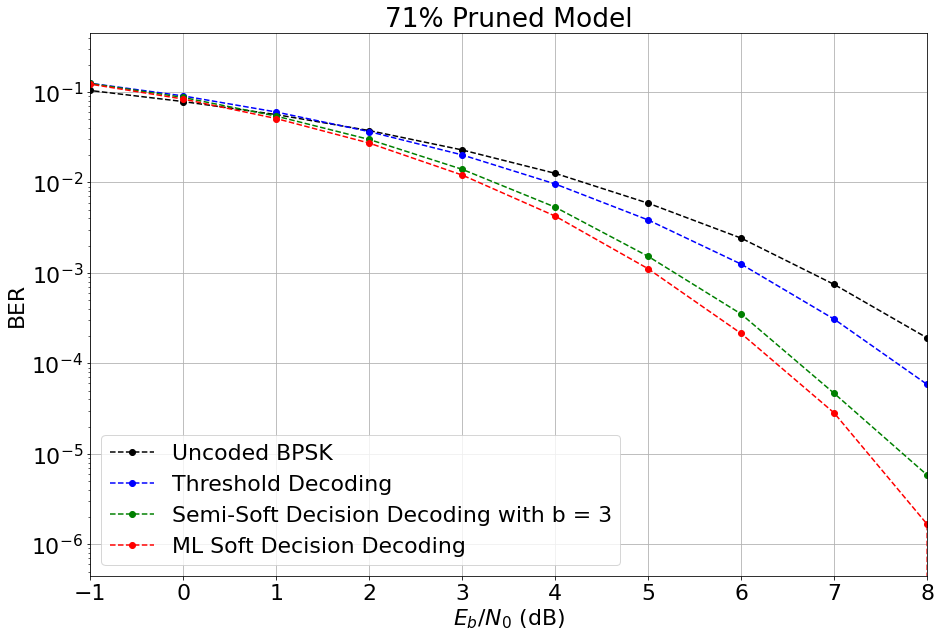}
    \label{fig:5d}
    } \quad
\caption{Performace Analysis of Neural Decoder for Polar (16,8) Codes with 2 hidden layers with $L_i$ = 256 for $i \in {1, 2}$. }
\label{fig:5}
\end{figure}


\subsection{Polar (16,8) Codes}

\textcolor{black}{We now test the LTH pruning approach with neural decoders for Polar (16,8) codes. Figure \ref{fig:5a} shows the observed test accuracies at 2 dB SNR, in response to varying levels of network pruning. Like before, we see that LTH retains the test accuracy of the original un-pruned network for a very wide range of pruning levels, and significantly outperforms standard pruning approaches. The BER curves of the various pruned network configurations are shown in Figure \ref{fig:5b}. In this case, we find that slight differences in the test accuracy at 2 dB SNR get amplified for higher SNR levels. For example, the 93\% pruned network is found to have only slightly lower test accuracy than less pruned configurations in Figure \ref{fig:5a}, but yields significantly worse BER in Figure \ref{fig:5b}.
} 

\textcolor{black}{Next, we test the impact of the proposed semi-soft decoding approach on the BER performances of the pruned networks. Results are shown in Figures \ref{fig:5c} and \ref{fig:5d}. Like before, we find that semi-soft decoding significantly enhances BER performances for the pruned neural decoders, for $b=2$ and $b=3$, enabling near optimal decoding.}

\section{Discussions and Conclusions}

\textcolor{black}{Our work establishes that iterative pruning approaches motivated by the lottery ticket hypothesis can reduce the complexity of neural decoders manyfold (up to 97\%), while maintaining decoding performance. This is verified across multiple experiments conducted with Hamming (7,4) and Polar (16,8) codes. We also study a \textit{semi-soft} decision decoding approach on the neural decoder outputs, which improves decoding performance in all cases.}

\textcolor{black}{Apart from the approaches demonstrated here, we are also investigating network weight quantization methods for reducing network complexity. We find that quantization can reduce network sizes up to a factor of 4, while maintaining performance. Through our initial investigative studies here, we hope to spark further research into combining pruning, semi-soft decoding and quantization based approaches for creating efficient low complexity neural decoders suitable for use in low-end communication edge devices. }

\enlargethispage{\baselineskip}

\newpage
\section*{Acknowledgments}

This work was supported in part by the Singapore Ministry of Education under Grant MOE2019-T2-2-171.




\bibliography{main}

\end{document}